\documentclass[conference]{IEEEtran}
\IEEEoverridecommandlockouts
\usepackage{cite}
\usepackage{amsmath,amssymb,amsfonts}
\usepackage[T1]{fontenc}
\usepackage[utf8]{inputenc}
\usepackage{algorithmic}
\usepackage{graphicx}
\usepackage{textcomp}
\usepackage{xcolor}
\def\BibTeX{{\rm B\kern-.05em{\sc i\kern-.025em b}\kern-.08em
    T\kern-.1667em\lower.7ex\hbox{E}\kern-.125emX}}

\usepackage{hyperref}

\usepackage{booktabs}
\usepackage{caption}
\usepackage[font=footnotesize]{subcaption}

\DeclareCaptionLabelSeparator{dot}{. }
\DeclareCaptionFormat{sbgamestable}{\fontsize{8pt}{8pt}\centering \rmfamily#1\break\scshape#3}
\DeclareCaptionFormat{sbgamessubtable}{\fontsize{8pt}{8pt}\centering\rmfamily#1\scshape#3}
\DeclareCaptionFormat{sbgamesfigure}{\fontsize{8pt}{8pt}\rmfamily#1#2#3}
\usepackage{tikz}
\usetikzlibrary{positioning}

\usepackage{pgfplots, pgfplotstable}
\pgfplotsset{compat=newest}

\pgfplotsset{
    show sum on top/.style={
        /pgfplots/scatter/@post marker code/.append code={%
            \node[
                at={(normalized axis cs:%
                        \pgfkeysvalueof{/data point/x},%
                        \pgfkeysvalueof{/data point/y})%
                },
                anchor=south,
            ]
            {\pgfmathprintnumber{\pgfkeysvalueof{/data point/y}}};
        },
    },
}

\newcommand\copyrighttext{%
  \footnotesize \textcopyright 2022 IEEE.  Personal use of this material is permitted.  Permission from IEEE must be obtained for all other uses, in any current or future media, including reprinting/republishing this material for advertising or promotional purposes, creating new collective works, for resale or redistribution to servers or lists, or reuse of any copyrighted component of this work in other works.}
\newcommand\copyrightnotice{%
\begin{tikzpicture}[remember picture,overlay]
\node[anchor=south,yshift=10pt] at (current page.south) {\fbox{\parbox{\dimexpr\textwidth-\fboxsep-\fboxrule\relax}{\copyrighttext}}};
\end{tikzpicture}%
}

\begin{document}

\title{Illuminating the Space of Enemies\\Through MAP-Elites}

\author{
\IEEEauthorblockN{Breno M. F. Viana}
\IEEEauthorblockA{\textit{Instituto de Ciências Matemáticas}\\\textit{e de Computação (ICMC)}\\
\textit{Universidade de São Paulo (USP)}\\
São Carlos, São Paulo, Brazil\\
\url{bmfviana@gmail.com}}
\and
\IEEEauthorblockN{Leonardo T. Pereira}
\IEEEauthorblockA{\textit{Instituto de Ciências Matemáticas}\\\textit{e de Computação (ICMC)}\\
\textit{Universidade de São Paulo (USP)}\\
São Carlos, São Paulo, Brazil\\
\url{leonardop@usp.br}}
\and
\IEEEauthorblockN{Claudio F. M. Toledo}
\IEEEauthorblockA{\textit{Instituto de Ciências Matemáticas}\\\textit{e de Computação (ICMC)}\\
\textit{Universidade de São Paulo (USP)}\\
São Carlos, São Paulo, Brazil\\
\url{claudio@icmc.usp.br}}
}

\maketitle
\copyrightnotice

\begin{abstract}
Action-Adventure games have several challenges to overcome, where the most common are enemies. The enemies' goal is to hinder the players' progression by taking life points, and the way they hinder this progress is distinct for different kinds of enemies. In this context, this paper introduces an extended version of an evolutionary approach for procedurally generating enemies that target the enemy's difficulty as the goal. Our approach advances the enemy generation research by incorporating a MAP-Elites population to generate diverse enemies without losing quality. The computational experiment showed the method converged most enemies in the MAP-Elites in less than a second for most cases. Besides, we experimented with players who played an Action-Adventure game prototype with enemies we generated. This experiment showed that the players enjoyed most levels they played, and we successfully created enemies perceived as easy, medium, or hard to face.
\end{abstract}


\section{Introduction} \label{sec:intro}

Enemies are crucial features of several game genres, such as Action-Adventure games. They offer challenges to the player since their goal is to hinder the player's progression and may require specific strategies to be defeated \cite{ref:viana2021procedural}. The enemies may provide different gameplay experiences, which is a relevant aspect for PCG methods \cite{ref:togelius2016introduction}. However,  PCG methods usually approach enemies by spreading them through a level map  \cite{ref:baldwin2017mixed, ref:baldwin2017towards, ref:liapis2017multi}, instead of concerning more about how to create themselves \cite{ref:khalifa2018talakat, ref:pereira2021procedural}.
The generation of enemies in terms of attributes and visuals design appeared in some games such as Spore \cite{ref:spore}, No Man's Sky \cite{ref:nomanssky}, and Creatures \cite{ref:creatures}.

We present in this paper an extension of the enemy generation algorithm introduced by Pereira et al. \cite{ref:pereira2021procedural}.
They introduced a Parallel Evolutionary Algorithm (PAE) to evolve enemies represented by common features of Action-Adventure games \cite{ref:pereira2021procedural}.
Our approach advances the evolutionary strategy by applying MAP-Elites population to illuminate enemies' space, and we did not evolve the enemies through parallel evolution.
Our MAP-Elites approach discretizes the enemy search space into a two-dimensional matrix mapped in terms of movement and weapon types.

The computational experiment showed that our approach converged most enemies in the MAP-Elites population with a generation process that took less than a second for all cases. We report results with players who enjoyed most of the gameplay, where enemies are successfully created based on easy, medium, or hard-to-face difficulty levels.

This paper is structured as follows.
\autoref{sec:relatedwork} presents the related works, and we highlighted our contributions.
\autoref{sec:methodology} describes the representation of our enemies and our evolutionary enemy generation approach.
\autoref{sec:results} presents and discusses the results of our experiments.
Finally, \autoref{sec:conclusion} presents the conclusions and future works.

\section{Related Works} \label{sec:relatedwork}

\begin{table*}[!ht]
\centering
\caption{Enemy generation literature summarizing and comparison with this work. `P' defines the partially generated enemy features.}
\label{table:enemygenerationworks}
\small
\begin{tabular}{@{}p{6cm}cccccc@{}}
\toprule
\textbf{Work} & \textbf{Placement} & \textbf{Amount} & \textbf{Status} & \textbf{Visuals} & \textbf{Adaptive} & \textbf{MAP-Elites} \\
\midrule
Baldwin et al. \cite{ref:baldwin2017mixed} & $\checkmark$ & $\checkmark$ & - & - & - & - \\
Baldwin et al. \cite{ref:baldwin2017towards} & $\checkmark$ & $\checkmark$ & - & - & - & - \\
Liapis \cite{ref:liapis2017multi} & $\checkmark$ & - & - & - & - & - \\
Khalifa et al. \cite{ref:khalifa2018talakat} & $\checkmark$ & $\checkmark$ & P & P & - & $\checkmark$ \\
Pereira et al. \cite{ref:pereira2021procedural} & $\checkmark$ & $\checkmark$ & $\checkmark$ & P & - & - \\
\midrule
Spore \cite{ref:spore} & - & - & P & P & - & - \\
No Man's Sky \cite{ref:nomanssky} & - & - & P & P & - & - \\
Creatures \cite{ref:creatures} & - & - & $\checkmark$ & P & $\checkmark$ & - \\
Diablo 3 \cite{ref:diabloiii} & - & - & P & - & - & - \\
Middle Earth: Shadow of Mordor \cite{ref:shadowofmordor} & - & - & P & - & - & - \\
Left 4 Dead 2 \cite{ref:left4dead2} & $\checkmark$ & $\checkmark$ & - & - & $\checkmark$ & - \\
State of Decay 2 \cite{ref:stateofdecay2} & $\checkmark$ & $\checkmark$ & - & - & $\checkmark$ & - \\
\midrule
This work & $\checkmark$ & - & $\checkmark$ & P & - & $\checkmark$ \\
\bottomrule
\end{tabular}
\end{table*}

The research on procedural enemy generation is a recent research area, \cite{ref:pereira2021procedural} and this section describes enemy generation methods from research works and industry games regarding any stage of their generative processes and gameplay progress.
\autoref{table:enemygenerationworks} summarizes our related work findings with the present paper contributions.
In this table, placement refers to the works that perform enemy placement. The amount refers to those works that control the number of enemies to place, while status relates to papers that generated the enemies' attributes. Visuals refer to those developing visual components of enemies, and adaptive refers to works applying adaptive generated for such contents. Finally, MAP-Elites refers to those using such a technique to create enemies.

Balwin et al. \cite{ref:baldwin2017mixed} developed the Evolutionary Dungeon Designer (EDD) to assist game designers in their creative process. The EDD generates dungeon levels with enemies and rewards through an evolutionary approach. The levels match micro-patterns regarding levels' structure defined by users as input.
The same authors later extended their approach to evolve meso-patterns that consider enemies and rewards besides the level structure \cite{ref:baldwin2017towards}. However, both algorithms only place enemies in their levels.

Liapis \cite{ref:liapis2017multi} introduced an evolutionary approach that works on two steps. First, the method evolves level sketches composed of walls and different rooms types to place them strategically.
Rooms are classified discretely with pre-defined numbers of rewards and enemies, and they may connect with other rooms in one, two, three, or eight directions. After dispersing rooms, the second stage evolves each room in a cavern-fashion way by placing walls and enemies strategically around rewards and protecting them from the player. Again, this work only puts enemies at their levels.

So far, the papers described only the placement of enemies instead of generating the enemies that players would face in their games.
Khalifa et al. \cite{ref:khalifa2018talakat} were pioneers in such an area by introducing an evolutionary approach to evolve Bullet Hell games' levels.
These games consist of bullets (enemies) with different damage values, speed, and movement patterns. To create the levels, they evolved Talakat scripts, a language proposed to describe their levels, through Constrained MAP-Elites (CME). These scripts have sections for spawners and the boss. The former section defines spawn points to spawn bullets or create new spawners, while the latter describes the boss's health, position, and behavior. The spawners have sets of parameters to determine the bullet it spawns, i.e., its speed, angle, size, and the angle rotation and speed of spawners. 

Therefore, they generate enemies that players face besides placing them in their levels. In a single execution, the authors developed a variety of levels without losing quality by using a Quality Diversity approach.
This class of algorithms is exciting for PCG purposes \cite{ref:gravina2019procedural}. More specifically, Khalifa et al. \cite{ref:khalifa2018talakat} applied an extension of the MAP-Elites, an Illumination Algorithm that returns a set of the best-found solutions discretized in a map regarding their features \cite{ref:mouret2015illuminating}. The original MAP-Elites is the one we applied in our work.

Pereira et al.\cite{ref:pereira2021procedural} presented a Parallel Evolutionary Algorithm (PEA) that generates enemies for an Action-Adventure game. The authors extracted the most common variables from enemies in different Action-Adventure games to build the enemy's genotype. Their PEA evolves enemies matching their difficulty degrees with the difficulty goal given as input. Their method inspires our approach, but the main differences are the MAP-Elites and the new difficulty function introduced here.


Regarding the industry games, the creation of Non-Playable Characters (NPCs) is present in \textit{Spore} \cite{ref:spore} and \textit{No Man's Sky} \cite{ref:nomanssky}. These NPCs may be confronted by players, just like enemies. The creatures of these games are created by randomly assigning different body parts. Their algorithms have some constraints for body parts, and they do not put together parts that cannot match another already selected. This strategy ensures the feasibility of procedural animations of the NPCs.

An older game series used an extended version of this concept, \textit{Creatures} \cite{ref:creatures} generated NPCs by evolving them physically and making them learn. These NPCs could learn about the environment, and the player's actions via a neural network that receives simulated senses using semi-symbolic approximation techniques as input \cite{ref:grand1998creatures}.

Regarding the generation of actual enemies, i.e., NPCs that actively look for fighting players, \textit{Diablo 3} \cite{ref:diabloiii} and \textit{Shadow of Mordor} \cite{ref:shadowofmordor} generated their enemies by changing some predefined characteristics. This behavior change approach allowed these games to make more diverse and unique challenges. In \textit{Left 4 Dead 2} \cite{ref:left4dead2} and \textit{State of Decay 2} \cite{ref:stateofdecay2}, the enemies can adapt to the players. However, instead of changing their features, like in the previous two games, \textit{Left 4 Dead 2} \cite{ref:left4dead2} and \textit{State of Decay 2} \cite{ref:stateofdecay2} decide how many and where to place enemies. 
They make such decisions accordingly to the players' performance.
If the player is doing well, new enemies are spawned, else the games spawn fewer enemies\footnote{The AI Systems of Left 4 Dead (\url{https://steamcdn-a.akamaihd.net/apps/valve/2009/ai_systems_of_l4d_mike_booth.pdf}).}
\footnote{Procedurally generating enemies, places, and loot in State of Decay 2 (\url{https://www.gamedeveloper.com/design/procedurally-generating-enemies-places-and-loot-in-i-state-of-decay-2-i-}).}.

\section{Methodology} \label{sec:methodology}

This section describes our enemy representation and how we evolve them through our MAP-Elites approach.

\subsection{Representation}

\begin{table*}[!ht]
\centering
\caption{List of attributes of the enemy’s genotype. The line between attributes represents the crossover point. Adapted from \cite{ref:pereira2021procedural}.}
\label{table:enemygenotype}
\small
\begin{tabular}{@{}lccp{8.5cm}@{}}
\toprule
\textbf{Attribute} & \textbf{Type} & \textbf{Range} & \textbf{Details (the attribute defines...)} \\
\midrule
Health & Integer & 1-5 & How many hits an enemy endures. \\
Damage & Integer & 1-4 & How many life points an enemy takes from the player. \\
Attack Speed & Float & 0.75-4.0 & How frequent an enemy shoots a projectile (1/Attack Speed). \\
Movement Type & Nominal & - & How the enemy moves during gameplay. \\
Movement Speed & Float & 0.8-2.8 & How faster the enemy moves. \\
Active Time & Float & 1.5-10.0 & The time in seconds that the enemy moves before resting. \\
Rest Time & Float & 0.3-1.5 & The time in seconds that the enemy rests before moving. \\
\midrule
Weapon Type & Nominal & - & The weapon gameplay properties. \\
Projectile Speed & Float & 1.0-4.0 & How faster the projectile moves towards the player. \\
\bottomrule
\end{tabular}
\end{table*}

As mentioned, we propose an evolutionary approach that advances from the method introduced in  \cite{ref:pereira2021procedural}, where our enemy genotypes come from the one presented in the previous work.
Such representation extracts common attributes of enemies in Action-Adventure games to build the enemy's genotype. These attributes are the following: health, damage, attack speed, movement speed, active time, rest time, movement type, weapon type, and projectile speed.
\autoref{table:enemygenotype} describes these attributes and shows their respective ranges of values. Our only change in the numerical values was the max movement speed value; we decreased it slightly because the max value was too fast.

Again in \autoref{table:enemygenotype}, movement type and weapon type are nominal attributes representing more complex behaviors and objects that enemies may have. Following, we list the types of movements:
\begin{itemize}
\item \textbf{None} the enemy stays still.
\item \textbf{Random} the enemy's movement is defined by a random direction 2D vector.
\item \textbf{Random 1D} the enemy's movement is determined by a random direction 1D vector (i.e., horizontal or vertical).
\item \textbf{Flee} the enemy's movement is calculated by the opposite of the player's direction vector.
\item \textbf{Flee 1D} the enemy's movement is calculated by the opposite of a single axe of the player's direction vector (i.e., horizontal or vertical).
\item \textbf{Follow} the enemy's movement is determined by the direction vector that points towards the player.
\item \textbf{Follow 1D} the enemy's movement is defined by a single axe of the direction vector that points towards the player (i.e., horizontal or vertical).
\end{itemize}
All these movements occur during the active time.
Regarding weapon types, we list their types and describe how they work:
\begin{itemize}
\item \textbf{Barehand (None)} deals damage on contact.
\item \textbf{Sword} deals damage on contact with a higher reach regarding the barehand.
\item \textbf{Bow} shoots bullets towards the player, and they deal damage when hit the player.
\item \textbf{Bomb-Thrower} shoots a bomb towards the player; they explode in 2 seconds and deal damage in a limited area.
\item \textbf{Shield} protects the enemy from frontal attacks.
\item \textbf{Cure Spell} cures one health point of all enemies in a circular area.
\end{itemize}

We add the cure spell to generate healer enemies, and our melee enemies use the following weapons: barehand, sword, and shield. Furthermore, our ranged enemies use a bow and bomb-thrower. We discarded the weights of the movement and weapon types and dealt with these attributes in dedicated equations to calculate the enemies' difficulty.

\subsection{Generation Process}

\begin{figure}[!t]
    \centering
    \begin{tikzpicture}[every node/.style={scale=0.6}]
    \tikzstyle{node} = [draw, minimum size=1cm]

    \node[node] (l1c1) at (0,0) { };
    \node[node] (l1c2) [right = 0.5cm of l1c1] { };
    \node[node] (l1c3) [right = 0.5cm of l1c2] { };
    \node[node] (l1c4) [right = 0.5cm of l1c3] { };
    \node[node] (l1c5) [right = 0.5cm of l1c4] { };
    \node[node] (l1c6) [right = 0.5cm of l1c5] { };
    \node[node] (l2c1) [below = 0.1cm of l1c1] { };
    \node[node] (l2c2) [right = 0.5cm of l2c1] { };
    \node[node] (l2c3) [right = 0.5cm of l2c2] { };
    \node[node] (l2c4) [right = 0.5cm of l2c3] { };
    \node[node] (l2c5) [right = 0.5cm of l2c4] { };
    \node[node] (l2c6) [right = 0.5cm of l2c5] { };
    \node[node] (l3c1) [below = 0.1cm of l2c1] { };
    \node[node] (l3c2) [right = 0.5cm of l3c1] { };
    \node[node] (l3c3) [right = 0.5cm of l3c2] { };
    \node[node] (l3c4) [right = 0.5cm of l3c3] { };
    \node[node] (l3c5) [right = 0.5cm of l3c4] { };
    \node[node] (l3c6) [right = 0.5cm of l3c5] { };
    \node[node] (l4c1) [below = 0.1cm of l3c1] { };
    \node[node] (l4c2) [right = 0.5cm of l4c1] { };
    \node[node] (l4c3) [right = 0.5cm of l4c2] { };
    \node[node, thick, blue!70!black] (l4c4) [right = 0.5cm of l4c3] { };
    \node[node] (l4c5) [right = 0.5cm of l4c4] { };
    \node[node] (l4c6) [right = 0.5cm of l4c5] { };
    \node[node] (l5c1) [below = 0.1cm of l4c1] { };
    \node[node] (l5c2) [right = 0.5cm of l5c1] { };
    \node[node] (l5c3) [right = 0.5cm of l5c2] { };
    \node[node] (l5c4) [right = 0.5cm of l5c3] { };
    \node[node] (l5c5) [right = 0.5cm of l5c4] { };
    \node[node] (l5c6) [right = 0.5cm of l5c5] { };
    \node[node] (l6c1) [below = 0.1cm of l5c1] { };
    \node[node, thick, red!70!black] (l6c2) [right = 0.5cm of l6c1] { };
    \node[node] (l6c3) [right = 0.5cm of l6c2] { };
    \node[node] (l6c4) [right = 0.5cm of l6c3] { };
    \node[node] (l6c5) [right = 0.5cm of l6c4] { };
    \node[node] (l6c6) [right = 0.5cm of l6c5] { };
    \node[node] (l7c1) [below = 0.1cm of l6c1] { };
    \node[node] (l7c2) [right = 0.5cm of l7c1] { };
    \node[node] (l7c3) [right = 0.5cm of l7c2] { };
    \node[node] (l7c4) [right = 0.5cm of l7c3] { };
    \node[node] (l7c5) [right = 0.5cm of l7c4] { };
    \node[node] (l7c6) [right = 0.5cm of l7c5] { };
    
    \node (ax) [below left = 0.15cm of l7c1] { };
    \node (ax_y1) [above left = 0.1cm and 0.1cm of l1c1] { };
    \node (ax_y2) [below right = 0.05cm and 0.2cm of l7c6] { };

    \draw[->] (ax.center) -- (ax_y1.center);
    \path (ax) -- node [rotate=90,below left=-0.45cm and 1.7cm] {\textbf{Movement Type}} (ax_y1);
    \node (le1) [left = 0.3cm of l1c1] {None};
    \node (le2) [left = 0.3cm of l2c1] {Random};
    \node (le3) [left = 0.3cm of l3c1] {Random 1D};
    \node[thick, blue!70!black] (le4) [left = 0.3cm of l4c1] {Flee};
    \node (le5) [left = 0.3cm of l5c1] {Flee 1D};
    \node[thick, red!70!black] (le6) [left = 0.3cm of l6c1] {Follow};
    \node (le7) [left = 0.3cm of l7c1] {Follow 1D};

    \draw[->] (ax.center) -- (ax_y2.center);
    \path (ax) -- node [below right=0.5cm and -1cm] {\textbf{Weapon Type}} (ax_y2);
    \node (ce1) [below = 0.3cm of l7c1] {Barehand};
    \node[thick, red!70!black] (ce2) [below = 0.3cm of l7c2] {Sword};
    \node (ce3) [below = 0.3cm of l7c3] {Bow};
    \node[thick, blue!70!black] (ce4) [below = 0.3cm of l7c4] {Bomb Thrower};
    \node (ce5) [below = 0.3cm of l7c5] {Shield};
    \node (ce6) [below = 0.3cm of l7c6] {Cure Spell};

    \draw[dashed, thick, red!70!black] (le6) -- (l6c2);
    \draw[dashed, thick, red!70!black] (ce2) -- (l6c2);
    
    \draw[dashed, thick, blue!70!black] (le4) -- (l4c4);
    \draw[dashed, thick, blue!70!black] (ce4) -- (l4c4);
    \end{tikzpicture}

    \caption{The map of MAP-Elites population. The red cell represents a melee enemy that follows the player to hit with a sword. The blue cell represents a ranged enemy that flees from the player while throwing bombs towards them.}
    \label{fig:map-representation}
\end{figure}
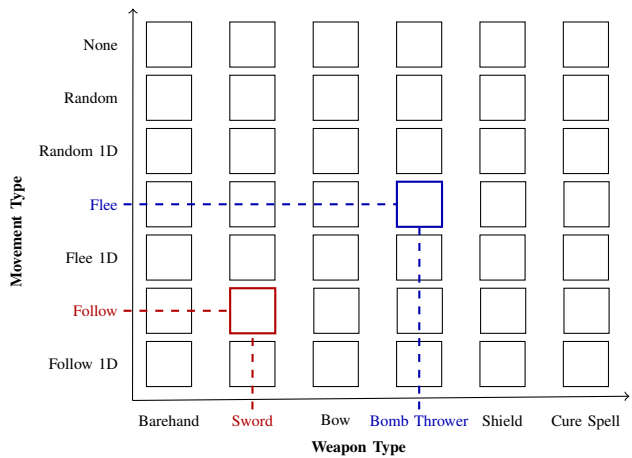

The input for the generation process is only the goal difficulty of enemies.
The fitness function measures the distance between aimed difficulty and the difficulty encoded in the enemy stated as an individual (representation of solution) of the evolutionary algorithm.
Therefore, our approach minimizes such fitness.
We designed a MAP-Elites approach to preserve diversity while optimizing the quality of enemies.
We discretized our map regarding movement and weapon types; thus, we have nominal values as feature descriptors (dimensions).
Since we do not need to calculate numerical equations, our mapping functions are straightforward:
\begin{equation}
    D_{\text{movement}} = e_{\mathit{movement\_type}}
\end{equation}
\begin{equation}
    D_{\mathit{weapon}} = e_{\mathit{weapon\_type}}
\end{equation}
where, $e$ is the enemy.
\autoref{fig:map-representation} presents our approach's map.
The cell highlighted in red represents an enemy that follows the player to hit with a sword, while in blue represents an enemy that flees from the player while throwing bombs towards them.

The proposed MAP-Elites approach maps 42 enemies in terms of the defined dimensions.
When the population receives a new individual, we calculate its feature descriptors to place it in the correct map entry.
If a new enemy hits a filled cell, we apply elitism, i.e., the best enemy fills the cell, discarding the other one.

The evolutionary process starts by generating the initial population thought filling the attributes of $n$ enemies randomly. Since the initialized enemies may hit the same map entry, the initial population generation may take a while. Besides, since we discretized the population in a map, its size does not change. Thus, the best individual is always kept.

Next, we evolve the population using the generation-limit stopping criterion. The authors in \cite{ref:pereira2021procedural} replace all the population in each generation by the intermediate population generated. We also have an intermediate population; however, we try to add its individuals in the MAP-Elites population. The reproduction operators create new individuals from two parents, chosen using tournament selection with two competitors.

We first perform a crossover with a 100\% rate to generate two new individuals when reproducing enemies. Our crossover is a combination of a fixed-single-point crossover and a BLX-$\alpha$ crossover \cite{ref:eshelman1993real}. We first cross the parents in the fixed point as shown in \autoref{table:enemygenotype}. We designed the crossover to fill our map faster once new individuals may hit new cells. For instance, if the elites Bow-Flee and Sword-Follow cross, we generate two individuals mapped in Bow-Follow and Sword-Flee cells.
After this, we perform the BLX-$\alpha$ crossover for each numerical attribute \cite{ref:eshelman1993real}.

After the crossover, we have a chance to mutate both resulting enemies and, when a mutation happens, we apply a multi-gene mutation \cite{ref:kanagal2014next}. To do so, we calculate the chance of mutating each gene. This mutation means that our mutation operator can change all the enemy's attributes. We set a new random value for each gene that mutates, respecting the limited range and the list of nominal values of the attributes.

Our difficulty function has four factors: health, movement, strength, and gameplay. The enemies' life points determine how many hits they endure.
\begin{equation}
d_{\mathit{health}} = 2 \times e_{\mathit{health}}
\end{equation}

Regarding the movement factor, we consider three attributes of our individuals: movement speed, active time, and rest time.
\begin{equation}
d_{\mathit{movement}} = e_{\mathit{mv\_spd}} + e_{\mathit{act\_tm}} / 3 + 1 / e_{\mathit{rst\_tm}}
\end{equation}
where, $\mathit{mv\_spd}$ is the movement speed, $\mathit{act\_tm}$ is the active time, and  $\mathit{rst\_tm}$ is the rest time.
The faster the enemy, the more difficult it will be for the player to defend the enemy's tackles. The more time active moving, the more difficult the enemy is. We weighed this term with $1/3$ to balance its influence in this equation. Finally, the more time resting, the more easily it will be to defeat; thus, we calculate its inverse.

The strength factor is more complex than the previous difficulty factors; it depends on the types of enemies and, therefore, we multiply three different equations.
\begin{equation}
d_{\mathit{strength}} = d_{\mathit{s_1}} \times d_{\mathit{s_2}} \times d_{\mathit{s_3}}
\end{equation}
For melee enemies, we multiply damage by movement speed.
\begin{equation}
d_{\mathit{s_1}} = \begin{cases}
            e_{\mathit{dmg}} \times e_{\mathit{mv\_spd}},& \textsc{isMelee}(e)\\
            1,& \text{otherwise}
          \end{cases}
\end{equation}
where, $dmg$ is the damage. We multiply attack speed by projectile speed for ranged enemies and weigh the result by three.
\begin{equation}
d_{\mathit{s_2}} = \begin{cases}
            3 \times (e_{\mathit{atk\_spd}} \times e_{\mathit{prjct\_spd}}),& \textsc{isRanged}(e)\\
            1,& \text{otherwise}
          \end{cases}
\end{equation}
where, $\mathit{atk\_spd}$ is the attack speed, and $\mathit{prjct\_spd}$ is the projectile speed.
We consider only the attack speed for healer enemies since they always heal a single life point of all enemies in their heal area range.
\begin{equation}
d_{\mathit{s_3}} = \begin{cases}
            2 \times e_{\mathit{atk\_spd}},& \textsc{isHealer}(e)\\
            1,& \text{otherwise}
          \end{cases}
\end{equation}

We also calculate the gameplay factor considering the enemies' weapons and our game prototype, where we experimented with the generated enemies. Here we also increase the difficulty of incoherent enemies; thus, discarding them based on a threshold defined by the user.
\begin{equation}
d_{\mathit{gameplay}} = d_{\mathit{g_1}} \times d_{\mathit{g_2}} \times d_{\mathit{g_3}} \times d_{\mathit{g_4}} \times d_{\mathit{g_5}}
\end{equation}

Melee enemies that follow the player are more dangerous, thus, more challenging to defeat. Moreover, melee enemies that flee from the player or stay still are less risky and easier to defeat. Therefore, we weigh the difficulty as follows.
\begin{equation}
d_{\mathit{g_1}} = \begin{cases}
            1.25, & \textsc{isMelee}(e) ~\text{and}~ \textsc{isFollow}(e)\\
            0.5, & \textsc{isMelee}(e) ~\text{and}~ \\
                & (\textsc{isAnyFlee}(e) ~\text{or}~ \textsc{hasNoMove}(e))\\
            1,& \text{otherwise}
          \end{cases}
\end{equation}

Since ranged enemies perform distance attacks, those that flee from the player present more risk. When ranged enemies stay still, players can defeat them easier since they are static targets. Ranged enemies that follow the player may have the projectile speed faster than their movement speed, or else they will not behave as rangers since their projectiles will be slower than they own. This behavior did not occur in enemies with the movement Follow1D. Thus, we weigh the difficulty as follows.
\begin{equation}
d_{\mathit{g_2}} = \begin{cases}
            1.25, & \textsc{isRanged}(e) ~\text{and}~ \textsc{isFlee}(e)\\
            1.15, & \textsc{isRanged}(e) ~\text{and}~ \textsc{isFlee1D}(e)\\
            0.5, & \textsc{isRanged}(e) ~\text{and}~ \textsc{hasNoMove}(e)\\
            1,& \text{otherwise}
          \end{cases}
\end{equation}
\begin{equation}
d_{\mathit{g_3}} = \begin{cases}
            0.5 / (2 \times e_{mv\_spd}), & \textsc{isRanged}(e) ~\text{and}~ \\
                 & \textsc{isFollow}(e)\\
            1,& \text{otherwise}
          \end{cases}
\end{equation}

Healers must protect themselves while keeping healing other enemies.
Thus, they should not follow players but avoid them.
Besides, healers that move faster are more difficult to defeat; thus, we also weighed this factor by their movement speed.
Therefore, we weigh the enemy's difficulty as follows.
\begin{equation}
d_{\mathit{g_4}} = \begin{cases}
            1, & \textsc{isHealer}(e) ~\text{and}~ \\
                 & (\textsc{isAnyRandom}(e) ~\text{or}~ \\
                 & \textsc{isAnyFlee}(e)) \\
            0.5,& \text{otherwise}
          \end{cases}
\end{equation}
\begin{equation}
d_{\mathit{g_5}} = \begin{cases}
            1.15 \times e_{mv\_spd}, & \textsc{isHealer}(e)\\
            1,& \text{otherwise}
          \end{cases}
\end{equation}

All the numeric weights in the equations were chosen empirically through gameplay experiments. Finally, we defined the final difficulty equation as follows.
\begin{equation}
d = d_{\mathit{gameplay}} \times (d_{\mathit{health}} + d_{\mathit{movement}} + d_{\mathit{strength}})
\end{equation}

\section{Results} \label{sec:results}

In this section, we present how our approach performed computationally and the feedback of human players after playing with our enemies.

\subsection{Performance Results}

\begin{table*}[!ht]
\caption{Results of fitness obtained after 100 executions of our approach. Each fitness value corresponds to the distance between the input difficulty and the difficulty of the found enemy -- the closer to zero, the better.}
\label{tab:fitness}

    \begin{subtable}[t]{\textwidth}
    \centering
    \caption{Very Easy Difficulty = 11.}
    \label{tab:difficulty11}
    \begin{tabular}{lcccccc}
    \toprule
                  & Barehand   & Sword      & Bow        & Bomb Thrower & Shield     & Cure Spell \\
    \midrule
    None     &  $0.00\pm0.00$ & $0.00\pm0.00$ & $0.01\pm0.01$ & $0.01\pm0.01$   & $0.00\pm0.00$ & $0.01\pm0.01$ \\
    Random   &  $0.00\pm0.00$ & $0.00\pm0.00$ & $0.01\pm0.01$ & $0.01\pm0.01$   & $0.00\pm0.00$ & $0.01\pm0.01$ \\
    Random1D &  $0.00\pm0.00$ & $0.00\pm0.00$ & $0.01\pm0.01$ & $0.01\pm0.01$   & $0.00\pm0.00$ & $0.01\pm0.01$ \\
    Flee     &  $0.00\pm0.00$ & $0.00\pm0.00$ & $0.16\pm0.23$ & $0.21\pm0.38$   & $0.00\pm0.00$ & $0.03\pm0.03$ \\
    Flee1D   &  $0.00\pm0.00$ & $0.00\pm0.00$ & $0.06\pm0.10$ & $0.07\pm0.12$   & $0.00\pm0.00$ & $0.03\pm0.04$ \\
    Follow   &  $0.01\pm0.01$ & $0.01\pm0.01$ & $0.43\pm0.58$ & $0.43\pm0.53$   & $0.01\pm0.02$ & $0.02\pm0.02$ \\
    Follow1D &  $0.00\pm0.00$ & $0.00\pm0.00$ & $0.01\pm0.01$ & $0.01\pm0.01$   & $0.00\pm0.00$ & $0.01\pm0.01$ \\
    \bottomrule
    \end{tabular}
    \end{subtable}

    \begin{subtable}[t]{\textwidth}
    \centering
    \caption{Easy Difficulty = 13.}
    \label{tab:difficulty13}
    \begin{tabular}{lcccccc}
    \toprule
                  & Barehand   & Sword      & Bow        & Bomb Thrower & Shield     & Cure Spell \\
    \midrule
    None     & $0.00\pm0.01$ & $0.00\pm0.00$ & $0.01\pm0.02$ & $0.01\pm0.02$   & $0.00\pm0.00$ & $0.01\pm0.01$ \\
    Random   & $0.00\pm0.00$ & $0.00\pm0.00$ & $0.01\pm0.01$ & $0.01\pm0.03$   & $0.00\pm0.00$ & $0.01\pm0.01$ \\
    Random1D & $0.00\pm0.00$ & $0.00\pm0.00$ & $0.01\pm0.01$ & $0.01\pm0.01$   & $0.00\pm0.00$ & $0.01\pm0.01$ \\
    Flee     & $0.00\pm0.02$ & $0.00\pm0.01$ & $0.12\pm0.22$ & $0.09\pm0.20$   & $0.00\pm0.01$ & $0.03\pm0.03$ \\
    Flee1D   & $0.00\pm0.01$ & $0.00\pm0.01$ & $0.04\pm0.05$ & $0.04\pm0.05$   & $0.00\pm0.01$ & $0.03\pm0.03$ \\
    Follow   & $0.01\pm0.01$ & $0.01\pm0.01$ & $1.07\pm1.12$ & $1.06\pm0.97$   & $0.01\pm0.01$ & $0.02\pm0.02$ \\
    Follow1D & $0.00\pm0.00$ & $0.00\pm0.00$ & $0.01\pm0.01$ & $0.01\pm0.01$   & $0.00\pm0.00$ & $0.01\pm0.01$ \\
    \bottomrule
    \end{tabular}
    \end{subtable}

    \begin{subtable}[t]{\textwidth}
    \centering
    \caption{Medium Difficulty = 15.}
    \label{tab:difficulty15}
    \begin{tabular}{lcccccc}
    \toprule
             & Barehand   & Sword      & Bow    & Bomb Thrower  & Shield     & Cure Spell \\
    \midrule
    None     & $0.07\pm0.14$ & $0.06\pm0.14$ & $0.02\pm0.03$ & $0.02\pm0.03$   & $0.08\pm0.18$ & $0.02\pm0.02$ \\
    Random   & $0.00\pm0.00$ & $0.00\pm0.00$ & $0.00\pm0.01$ & $0.00\pm0.01$   & $0.00\pm0.00$ & $0.01\pm0.01$ \\
    Random1D & $0.00\pm0.00$ & $0.00\pm0.00$ & $0.00\pm0.01$ & $0.00\pm0.01$   & $0.00\pm0.00$ & $0.01\pm0.02$ \\
    Flee     & $0.07\pm0.14$ & $0.08\pm0.23$ & $0.03\pm0.04$ & $0.03\pm0.04$   & $0.08\pm0.22$ & $0.03\pm0.03$ \\
    Flee1D   & $0.05\pm0.11$ & $0.06\pm0.13$ & $0.02\pm0.02$ & $0.02\pm0.03$   & $0.07\pm0.14$ & $0.02\pm0.03$ \\
    Follow   & $0.01\pm0.01$ & $0.01\pm0.01$ & $1.70\pm1.47$ & $2.15\pm1.82$   & $0.01\pm0.01$ & $0.02\pm0.02$ \\
    Follow1D & $0.00\pm0.00$ & $0.00\pm0.00$ & $0.00\pm0.01$ & $0.00\pm0.00$   & $0.00\pm0.00$ & $0.01\pm0.01$ \\
    \bottomrule
    \end{tabular}
    \end{subtable}

    \begin{subtable}[t]{\textwidth}
    \centering
    \caption{Hard Difficulty = 17.}
    \label{tab:difficulty17}
    \begin{tabular}{lcccccc}
    \toprule
             & Barehand   & Sword      & Bow    & Bomb Thrower  & Shield     & Cure Spell \\
    \midrule
    None     & $1.91\pm0.25$ & $1.90\pm0.20$ & $0.03\pm0.03$ & $0.03\pm0.03$   & $1.91\pm0.21$ & $0.02\pm0.02$ \\
    Random   & $0.00\pm0.00$ & $0.00\pm0.00$ & $0.00\pm0.00$ & $0.00\pm0.00$   & $0.00\pm0.00$ & $0.01\pm0.01$ \\
    Random1D & $0.00\pm0.00$ & $0.00\pm0.00$ & $0.00\pm0.00$ & $0.00\pm0.00$   & $0.00\pm0.00$ & $0.01\pm0.01$ \\
    Flee     & $1.96\pm0.27$ & $1.94\pm0.27$ & $0.02\pm0.03$ & $0.02\pm0.01$   & $1.96\pm0.27$ & $0.02\pm0.02$ \\
    Flee1D   & $1.98\pm0.27$ & $1.94\pm0.24$ & $0.01\pm0.01$ & $0.01\pm0.01$   & $1.95\pm0.26$ & $0.02\pm0.02$ \\
    Follow   & $0.01\pm0.01$ & $0.01\pm0.01$ & $2.32\pm2.07$ & $2.29\pm2.16$   & $0.01\pm0.01$ & $0.01\pm0.01$ \\
    Follow1D & $0.00\pm0.00$ & $0.00\pm0.00$ & $0.00\pm0.00$ & $0.00\pm0.00$   & $0.00\pm0.00$ & $0.01\pm0.01$ \\
    \bottomrule
    \end{tabular}
    \end{subtable}

    \begin{subtable}[t]{\textwidth}
    \centering
    \caption{Very Hard Difficulty = 19.}
    \label{tab:difficulty19}
    \begin{tabular}{lcccccc}
    \toprule
         & Barehand    & Sword       & Bow        & Bomb Thrower & Shield      & Cure Spell \\
    \midrule
    None     & $3.91\pm0.23$ & $3.93\pm0.24$ & $0.04\pm0.04$ & $0.03\pm0.04$   & $3.93\pm0.27$ & $0.02\pm0.02$ \\
    Random   & $0.00\pm0.00$ & $0.00\pm0.00$ & $0.00\pm0.00$ & $0.00\pm0.00$   & $0.00\pm0.00$ & $0.01\pm0.01$ \\
    Random1D & $0.00\pm0.00$ & $0.00\pm0.00$ & $0.00\pm0.00$ & $0.00\pm0.00$   & $0.00\pm0.00$ & $0.01\pm0.01$ \\
    Flee     & $3.95\pm0.27$ & $3.99\pm0.32$ & $0.02\pm0.02$ & $0.01\pm0.02$   & $3.94\pm0.29$ & $0.02\pm0.02$ \\
    Flee1D   & $3.90\pm0.18$ & $3.89\pm0.17$ & $0.01\pm0.01$ & $0.01\pm0.01$   & $3.91\pm0.18$ & $0.02\pm0.02$ \\
    Follow   & $0.01\pm0.01$ & $0.01\pm0.01$ & $3.14\pm2.39$ & $3.35\pm2.63$   & $0.01\pm0.01$ & $0.01\pm0.01$ \\
    Follow1D & $0.00\pm0.00$ & $0.00\pm0.00$ & $0.00\pm0.00$ & $0.00\pm0.00$   & $0.00\pm0.00$ & $0.01\pm0.01$ \\
    \bottomrule
    \end{tabular}
    \end{subtable}

\end{table*}

\begin{table}[!ht]
\centering
\caption{Results of time in seconds obtained after 100 executions of our approach.}
\label{tab:time}

\begin{tabular}{@{}ccccc@{}}
\toprule
Difficulty & Average & Minimum & Maximum & Standard Deviation \\
\midrule
11         & 0.1608  & 0.1510  & 0.2345  & 0.0126 \\
13         & 0.1609  & 0.1521  & 0.2255  & 0.0115 \\
15         & 0.1597  & 0.1508  & 0.2474  & 0.0145 \\
17         & 0.1621  & 0.1509  & 0.2000  & 0.0106 \\
19         & 0.1581  & 0.1512  & 0.1970  & 0.0072 \\
\bottomrule
\end{tabular}
\end{table}

We defined the parameters of our approach empirically after comparing some range of values. The results comparing different sets of evolutionary parameters are available in a Google Sheets spreadsheet\footnote{Link to the spreadsheet \href{https://docs.google.com/spreadsheets/d/19SMHZYT_pfniZDuNS0BOYMt1kWyBOlvqFq_Y7vBNrS4}{https://docs.google.com/spreadsheets/d/19SMHZ\break YT\_pfniZDuNS0BOYMt1kWyBOlvqFq\_Y7vBNrS4.}}.
After this comparison, we set the following parameters: 500 generations as stop-criterion, 35 individuals for initial population, 100 individuals for intermediate population, 20\% for mutation rate, 30\% for gene mutation rate, 2-size for tournament selection. Next, we collected data from 100 executions of our method for three different difficulty goals to evaluate its performance. In such an experiment, each player could play three different levels with easy, medium, and hard enemies, respectively. We opt for this to provide short gameplay sections. \autoref{tab:fitness} shows the average and standard deviation of the fitness for each Elite (entry) of our MAP-Elites population, where most solutions converged to zero, the best value when using a distance measure for fitness. 

We expect values close to zero for lower difficulty values because these are easier to reach. Still, in \autoref{tab:fitness}, Bow and Bomb Thrower enemies with movements of Flee and Follow presented higher values. We expect such values for ranged enemies with Flee movement, once this movement makes enemies more dangerous. Elite gets closer to zero in the remaining tables, meaning the respective Elite is reaching the aimed difficulty. On the other hand, regarding the ranged enemies with the movement Follow, their distance is a little high due to a lack of balance between projectile and movement speed attributes. This result is similar in the remaining cases.

In tables \ref{tab:difficulty17} and \ref{tab:difficulty19}, we observe that new Elites in both tables have significantly higher distance values than the others. These Elites are the enemies with the Barehand, Sword, and Shield as weapons and None, Flee, and Flee1D as movements. We expected such results because these values are the ones we set with lesser values for their weights in the gameplay factor of our difficulty function. Besides, we also observe a difficulty limit in these enemies for both tables. These values vary between 15 and 16 since the distance between them and the difficulties 17 and 19 are, respectively, 2 and 4 approximately.

Regarding the execution time, \autoref{tab:time} presents the average, minimum, maximum, and standard deviation values for duration time (seconds) for the 100 executions. The experiments regarding performance were carried out in a PC with the following setup: Intel Core i7-7700HQ 2.80GHz Processor (8 cores), 16 GB DDR4 RAM, 236GB SSD memory, NVIDIA GeForce GTX 1050 Ti 4GB graphics card. The results show that different values of difficulty goals do not impact the execution time of our approach.
Therefore, our approach can generate enemies of any difficulty without performance loss.

The Parallel Evolutionary Algorithm presented by Pereira et al. \cite{ref:pereira2021procedural} can generate a huge number of enemies in a single execution. Although our approach generates significantly fewer enemies, it ensures diversity. Our method was slightly faster regarding average execution time, since their loweest average time was 0.168 seconds.

\subsection{Gameplay Feedback}

\begin{figure}[!t]
\centerline{\includegraphics[width=.95\linewidth]{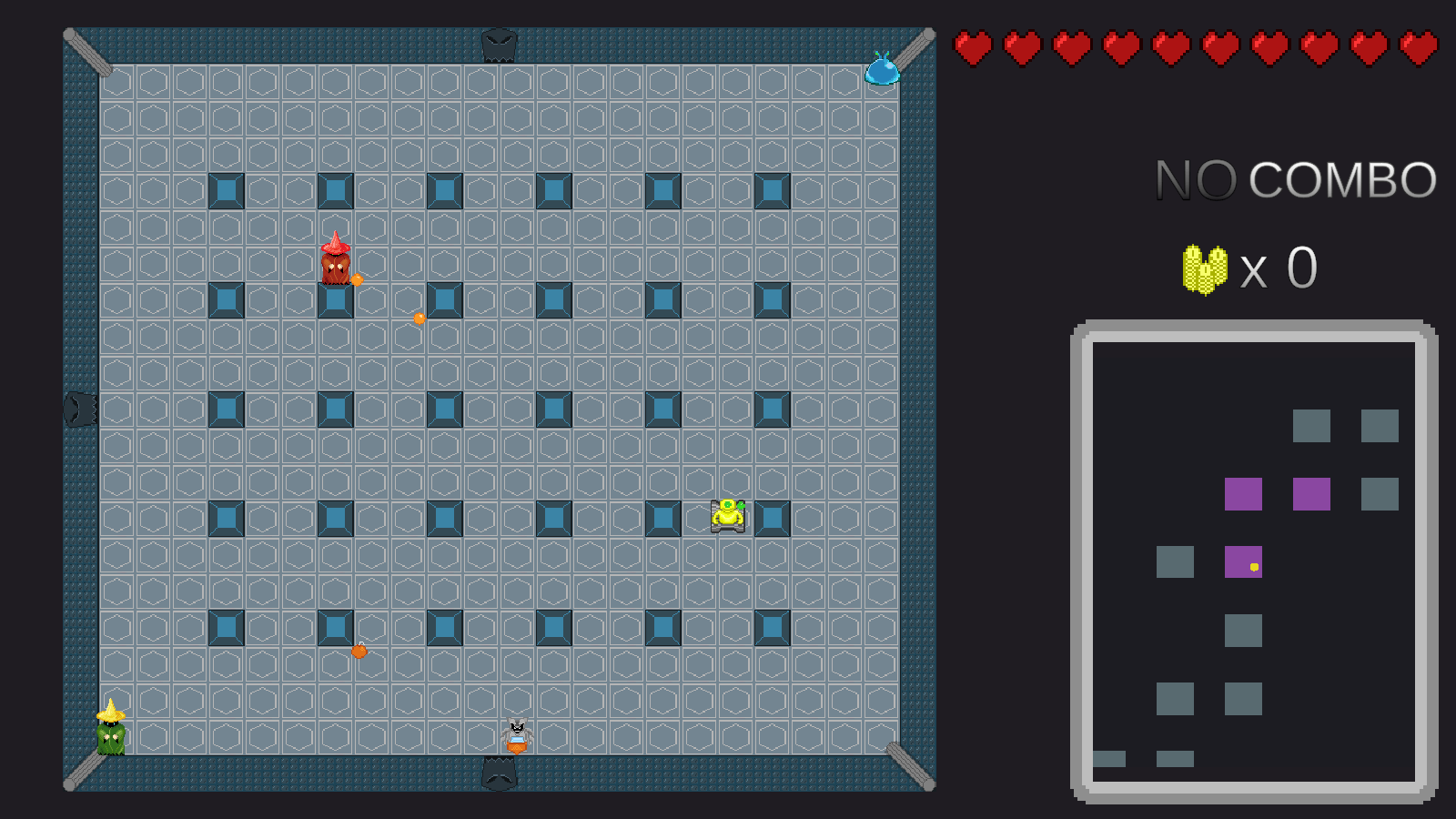}}
\caption{Game prototype screenshot.}
\label{fig:gameprototype}
\end{figure}

\begin{figure}[!t]
\centering

    \begin{subfigure}[t]{.15\textwidth}
    \centering
    \includegraphics[width=.35\linewidth]{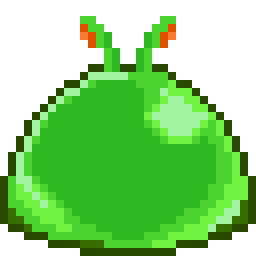}
    \subcaption{Slime (no weapon).}
    \label{fig:e1}
    \end{subfigure}
    ~
    \begin{subfigure}[t]{.15\textwidth}
    \centering
    \includegraphics[width=.5\linewidth]{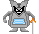}
    \subcaption{Swordsman.}
    \label{fig:e2}
    \end{subfigure}
    ~
    \begin{subfigure}[t]{.15\textwidth}
    \centering
    \includegraphics[width=.3\linewidth]{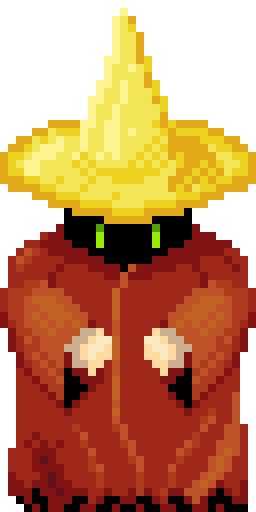}
    \subcaption{Bower mage.}
    \label{fig:e3}
    \end{subfigure}
    ~
    \begin{subfigure}[t]{.15\textwidth}
    \centering
    \includegraphics[width=.3\linewidth]{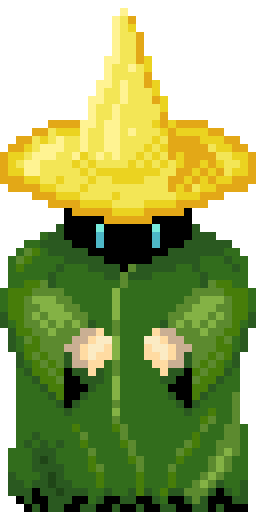}
    \subcaption{Bomber mage.}
    \label{fig:e4}
    \end{subfigure}
    ~
    \begin{subfigure}[t]{.15\textwidth}
    \centering
    \includegraphics[width=.5\linewidth]{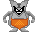}
    \subcaption{Shieldsman.}
    \label{fig:e5}
    \end{subfigure}
    ~
    \begin{subfigure}[t]{.15\textwidth}
    \centering
    \includegraphics[width=.3\linewidth]{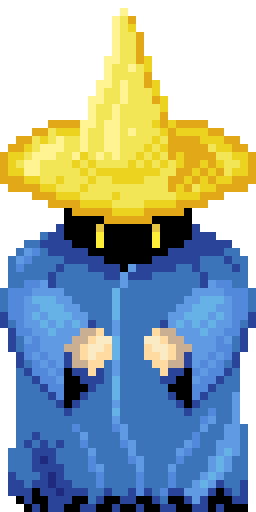}
    \subcaption{Healer (Cure Spell).}
    \label{fig:e6}
    \end{subfigure}

\caption{List of enemies of our game prototype. Slimes have no weapon. Swordsmans use swords. Bower mages shoot arrows. Bomber mages throw bombs. Shieldsmans hold shields. Healers use cure spell to heal other enemies.}
\label{fig:enemies}

\end{figure}

We experimented with anonymous volunteers who evaluated the quality of enemies through an in-game, optional, questionnaire. We shared the game, hosted in a server, via social media and mailing lists. Sets of enemies were generated with different difficulty levels and included in our game prototype, which is an adapted version of the game applied in \cite{ref:pereira2021procedural}. Therefore, the players must solve locked-door puzzles, defeat enemies and reach the level goal. The players can use blocks within some rooms to protect themselves from enemies, which are positioned in rooms like in  \cite{ref:pereira2021procedural}. \autoref{fig:gameprototype} shows a screenshot of such game prototype. In our game, players should defeat the six enemies listed in \autoref{fig:enemies}.

\begin{figure*}[!ht]
    \centering
    
    \begin{subfigure}[t]{.2\textwidth}
    \centering
    \begin{tikzpicture}[every node/.style={scale=0.8}]
        \begin{axis}
        [
            ylabel = Answers,
            xlabel = 5-point Likert Scale,
            ybar stacked,
            height=3.25cm,
            width=4.5cm,
            ymin = 0,
            ymax = 55,
            ytick={0,10,20,30,40,50},
            xtick={1,2,3,4,5},
            every axis plot/.append style={fill},
            nodes near coords={},
        ]
        \addplot [
            red!55!white!75!black,
            text=black,
            show sum on top,
        ] coordinates {
            (1,2)
            (2,18)
            (3,0)
            (4,0)
            (5,0)
        };
        \addplot [
            blue!55!white!75!black,
            text=black,
            show sum on top,
        ] coordinates {
            (1,0)
            (2,0)
            (3,22)
            (4,0)
            (5,0)
        };
        \addplot [
            green!55!white!75!black,
            text=black,
            show sum on top,
        ] coordinates {
            (1,0)
            (2,0)
            (3,0)
            (4,36)
            (5,46)
        };
        \end{axis}
    \end{tikzpicture}
    \subcaption{Q1 (Level was fun).}
    \label{fig:q1}
    \end{subfigure}
    ~
    \begin{subfigure}[t]{.22\textwidth}
    \centering
    \begin{tikzpicture}[every node/.style={scale=0.8}]
        \begin{axis}
        [
            ylabel = Answers,
            xlabel = 5-point Likert Scale,
            ybar stacked,
            height=3.25cm,
            width=4.5cm,
            ymin = 0,
            ymax = 55,
            ytick={0,10,20,30,40,50},
            xtick={1,2,3,4,5},
            every axis plot/.append style={fill},
            nodes near coords={},
        ]
        \addplot [
            red!55!white!75!black,
            text=black,
            show sum on top,
        ] coordinates {
            (1,25)
            (2,21)
            (3,0)
            (4,0)
            (5,0)
        };
        \addplot [
            blue!55!white!75!black,
            text=black,
            show sum on top,
        ] coordinates {
            (1,0)
            (2,0)
            (3,34)
            (4,0)
            (5,0)
        };
        \addplot [
            green!55!white!75!black,
            text=black,
            show sum on top,
        ] coordinates {
            (1,0)
            (2,0)
            (3,0)
            (4,24)
            (5,20)
        };
        \end{axis}
    \end{tikzpicture}
    \subcaption{Q2 (Enemies were difficult).}
    \label{fig:q2}
    \end{subfigure}
    ~
    \begin{subfigure}[t]{.2\textwidth}
    \centering
    \begin{tikzpicture}[every node/.style={scale=0.8}]
        \begin{axis}
        [
            ylabel = Answers,
            xlabel = 5-point Likert Scale,
            ybar stacked,
            height=3.25cm,
            width=4.5cm,
            ymin = 0,
            ymax = 55,
            ytick={0,10,20,30,40,50},
            xtick={1,2,3,4,5},
            every axis plot/.append style={fill},
            nodes near coords={},
        ]
        \addplot [
            red!55!white!75!black,
            text=black,
            show sum on top,
        ] coordinates {
            (1,15)
            (2,22)
            (3,0)
            (4,0)
            (5,0)
        };
        \addplot [
            blue!55!white!75!black,
            text=black,
            show sum on top,
        ] coordinates {
            (1,0)
            (2,0)
            (3,25)
            (4,0)
            (5,0)
        };
        \addplot [
            green!55!white!75!black,
            text=black,
            show sum on top,
        ] coordinates {
            (1,0)
            (2,0)
            (3,0)
            (4,39)
            (5,23)
        };
        \end{axis}
    \end{tikzpicture}
    \subcaption{Q3 (Balance was right).}
    \label{fig:q3}
    \end{subfigure}
    ~
    \begin{subfigure}[t]{.25\textwidth}
    \centering
    \begin{tikzpicture}[every node/.style={scale=0.8}]
        \begin{axis}
        [
            ylabel = Answers,
            xlabel = 5-point Likert Scale,
            ybar stacked,
            height=3.25cm,
            width=4.5cm,
            ymin = 0,
            ymax = 55,
            ytick={0,10,20,30,40,50},
            xtick={1,2,3,4,5},
            every axis plot/.append style={fill},
            nodes near coords={},
        ]
        \addplot [
            red!55!white!75!black,
            text=black,
            show sum on top,
        ] coordinates {
            (1,21)
            (2,20)
            (3,0)
            (4,0)
            (5,0)
        };
        \addplot [
            blue!55!white!75!black,
            text=black,
            show sum on top,
        ] coordinates {
            (1,0)
            (2,0)
            (3,25)
            (4,0)
            (5,0)
        };
        \addplot [
            green!55!white!75!black,
            text=black,
            show sum on top,
        ] coordinates {
            (1,0)
            (2,0)
            (3,0)
            (4,29)
            (5,29)
        };
        \end{axis}
    \end{tikzpicture}
    \subcaption{Q4 (Enemies were human-made).}
    \label{fig:q4}
    \end{subfigure}
    
    \caption{Bar charts of answers of the 75 players after playing 124 levels.}
    \label{fig:feedback}
\end{figure*}
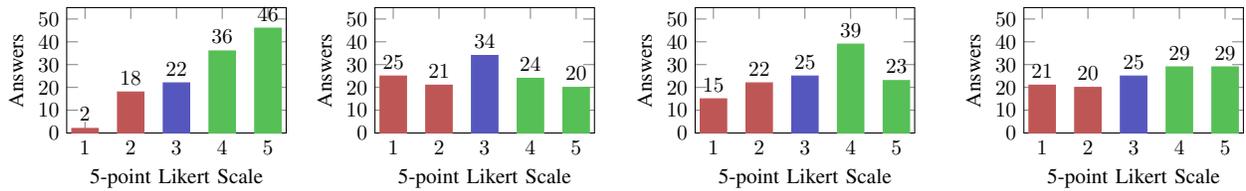

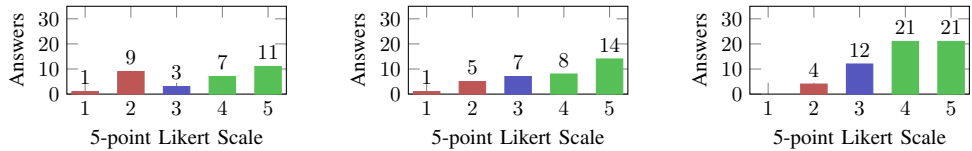
\begin{figure*}[!ht]
    \centering
    
    \begin{subfigure}[t]{.23\textwidth}
    \centering
    \begin{tikzpicture}[every node/.style={scale=0.8}]
        \begin{axis}
        [
            ylabel = Answers,
            xlabel = 5-point Likert Scale,
            ybar stacked,
            height=2.75cm,
            width=4.5cm,
            ymin = 0,
            ymax = 35,
            ytick={0,10,20,30},
            xtick={1,2,3,4,5},
            every axis plot/.append style={fill},
            nodes near coords={},
        ]
        \addplot [
            red!55!white!75!black,
            text=black,
            show sum on top,
        ] coordinates {
            (1,1)
            (2,9)
            (3,0)
            (4,0)
            (5,0)
        };
        \addplot [
            blue!55!white!75!black,
            text=black,
            show sum on top,
        ] coordinates {
            (1,0)
            (2,0)
            (3,3)
            (4,0)
            (5,0)
        };
        \addplot [
            green!55!white!75!black,
            text=black,
            show sum on top,
        ] coordinates {
            (1,0)
            (2,0)
            (3,0)
            (4,7)
            (5,11)
        };
        \end{axis}
    \end{tikzpicture}
    \subcaption{Easy.}
    \label{fig:q1easy}
    \end{subfigure}
    ~
    \begin{subfigure}[t]{.23\textwidth}
    \centering
    \begin{tikzpicture}[every node/.style={scale=0.8}]
        \begin{axis}
        [
            ylabel = Answers,
            xlabel = 5-point Likert Scale,
            ybar stacked,
            height=2.75cm,
            width=4.5cm,
            ymin = 0,
            ymax = 35,
            ytick={0,10,20,30},
            xtick={1,2,3,4,5},
            every axis plot/.append style={fill},
            nodes near coords={},
        ]
        \addplot [
            red!55!white!75!black,
            text=black,
            show sum on top,
        ] coordinates {
            (1,1)
            (2,5)
            (3,0)
            (4,0)
            (5,0)
        };
        \addplot [
            blue!55!white!75!black,
            text=black,
            show sum on top,
        ] coordinates {
            (1,0)
            (2,0)
            (3,7)
            (4,0)
            (5,0)
        };
        \addplot [
            green!55!white!75!black,
            text=black,
            show sum on top,
        ] coordinates {
            (1,0)
            (2,0)
            (3,0)
            (4,8)
            (5,14)
        };
        \end{axis}
    \end{tikzpicture}
    \subcaption{Medium.}
    \label{fig:q1medium}
    \end{subfigure}
    ~
    \begin{subfigure}[t]{.23\textwidth}
    \centering
    \begin{tikzpicture}[every node/.style={scale=0.8}]
        \begin{axis}
        [
            ylabel = Answers,
            xlabel = 5-point Likert Scale,
            ybar stacked,
            height=2.75cm,
            width=4.5cm,
            ymin = 0,
            ymax = 35,
            ytick={0,10,20,30},
            xtick={1,2,3,4,5},
            every axis plot/.append style={fill},
            nodes near coords={},
        ]
        \addplot [
            red!55!white!75!black,
            text=black,
            show sum on top,
        ] coordinates {
            (1,0)
            (2,4)
            (3,0)
            (4,0)
            (5,0)
        };
        \addplot [
            blue!55!white!75!black,
            text=black,
            show sum on top,
        ] coordinates {
            (1,0)
            (2,0)
            (3,12)
            (4,0)
            (5,0)
        };
        \addplot [
            green!55!white!75!black,
            text=black,
            show sum on top,
        ] coordinates {
            (1,0)
            (2,0)
            (3,0)
            (4,21)
            (5,21)
        };
        \end{axis}
    \end{tikzpicture}
    \subcaption{Hard.}
    \label{fig:q1hard}
    \end{subfigure}
    
    \caption{Bar charts of answers for question Q1 (``The level was fun to play'').}
    \label{fig:feedback_q1}
\end{figure*}

\begin{figure*}[!ht]
    \centering
    
    \begin{subfigure}[t]{.23\textwidth}
    \centering
    \begin{tikzpicture}[every node/.style={scale=0.8}]
        \begin{axis}
        [
            ylabel = Answers,
            xlabel = 5-point Likert Scale,
            ybar stacked,
            height=2.75cm,
            width=4.5cm,
            ymin = 0,
            ymax = 35,
            ytick={0,10,20,30},
            xtick={1,2,3,4,5},
            every axis plot/.append style={fill},
            nodes near coords={},
        ]
        \addplot [
            red!55!white!75!black,
            text=black,
            show sum on top,
        ] coordinates {
            (1,14)
            (2,6)
            (3,0)
            (4,0)
            (5,0)
        };
        \addplot [
            blue!55!white!75!black,
            text=black,
            show sum on top,
        ] coordinates {
            (1,0)
            (2,0)
            (3,7)
            (4,0)
            (5,0)
        };
        \addplot [
            green!55!white!75!black,
            text=black,
            show sum on top,
        ] coordinates {
            (1,0)
            (2,0)
            (3,0)
            (4,2)
            (5,2)
        };
        \end{axis}
    \end{tikzpicture}
    \subcaption{Easy.}
    \label{fig:q2easy}
    \end{subfigure}
    ~
    \begin{subfigure}[t]{.23\textwidth}
    \centering
    \begin{tikzpicture}[every node/.style={scale=0.8}]
        \begin{axis}
        [
            ylabel = Answers,
            xlabel = 5-point Likert Scale,
            ybar stacked,
            height=2.75cm,
            width=4.5cm,
            ymin = 0,
            ymax = 35,
            ytick={0,10,20,30},
            xtick={1,2,3,4,5},
            every axis plot/.append style={fill},
            nodes near coords={},
        ]
        \addplot [
            red!55!white!75!black,
            text=black,
            show sum on top,
        ] coordinates {
            (1,9)
            (2,6)
            (3,0)
            (4,0)
            (5,0)
        };
        \addplot [
            blue!55!white!75!black,
            text=black,
            show sum on top,
        ] coordinates {
            (1,0)
            (2,0)
            (3,9)
            (4,0)
            (5,0)
        };
        \addplot [
            green!55!white!75!black,
            text=black,
            show sum on top,
        ] coordinates {
            (1,0)
            (2,0)
            (3,0)
            (4,5)
            (5,6)
        };
        \end{axis}
    \end{tikzpicture}
    \subcaption{Medium.}
    \label{fig:q2medium}
    \end{subfigure}
    ~
    \begin{subfigure}[t]{.23\textwidth}
    \centering
    \begin{tikzpicture}[every node/.style={scale=0.8}]
        \begin{axis}
        [
            ylabel = Answers,
            xlabel = 5-point Likert Scale,
            ybar stacked,
            height=2.75cm,
            width=4.5cm,
            ymin = 0,
            ymax = 35,
            ytick={0,10,20,30},
            xtick={1,2,3,4,5},
            every axis plot/.append style={fill},
            nodes near coords={},
        ]
        \addplot [
            red!55!white!75!black,
            text=black,
            show sum on top,
        ] coordinates {
            (1,2)
            (2,9)
            (3,0)
            (4,0)
            (5,0)
        };
        \addplot [
            blue!55!white!75!black,
            text=black,
            show sum on top,
        ] coordinates {
            (1,0)
            (2,0)
            (3,18)
            (4,0)
            (5,0)
        };
        \addplot [
            green!55!white!75!black,
            text=black,
            show sum on top,
        ] coordinates {
            (1,0)
            (2,0)
            (3,0)
            (4,17)
            (5,12)
        };
        \end{axis}
    \end{tikzpicture}
    \subcaption{Hard.}
    \label{fig:q2hard}
    \end{subfigure}
    
    \caption{Bar charts of answers for question Q2 (``The enemies of this level were difficult to defeat'').}
    \label{fig:feedback_q2}
\end{figure*}
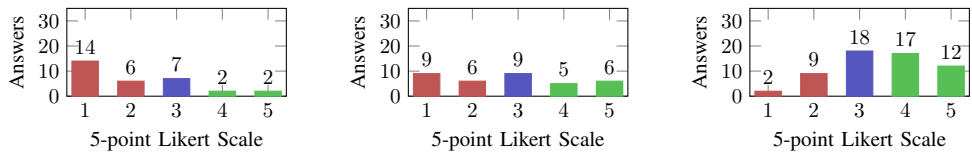

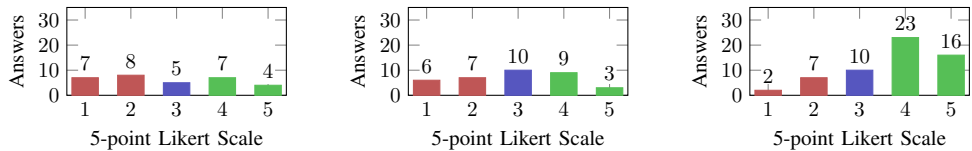
\begin{figure*}[!ht]
    \centering
    
    \begin{subfigure}[t]{.23\textwidth}
    \centering
    \begin{tikzpicture}[every node/.style={scale=0.8}]
        \begin{axis}
        [
            ylabel = Answers,
            xlabel = 5-point Likert Scale,
            ybar stacked,
            height=2.75cm,
            width=4.5cm,
            ymin = 0,
            ymax = 35,
            ytick={0,10,20,30},
            xtick={1,2,3,4,5},
            every axis plot/.append style={fill},
            nodes near coords={},
        ]
        \addplot [
            red!55!white!75!black,
            text=black,
            show sum on top,
        ] coordinates {
            (1,7)
            (2,8)
            (3,0)
            (4,0)
            (5,0)
        };
        \addplot [
            blue!55!white!75!black,
            text=black,
            show sum on top,
        ] coordinates {
            (1,0)
            (2,0)
            (3,5)
            (4,0)
            (5,0)
        };
        \addplot [
            green!55!white!75!black,
            text=black,
            show sum on top,
        ] coordinates {
            (1,0)
            (2,0)
            (3,0)
            (4,7)
            (5,4)
        };
        \end{axis}
    \end{tikzpicture}
    \subcaption{Easy.}
    \label{fig:q3easy}
    \end{subfigure}
    ~
    \begin{subfigure}[t]{.23\textwidth}
    \centering
    \begin{tikzpicture}[every node/.style={scale=0.8}]
        \begin{axis}
        [
            ylabel = Answers,
            xlabel = 5-point Likert Scale,
            ybar stacked,
            height=2.75cm,
            width=4.5cm,
            ymin = 0,
            ymax = 35,
            ytick={0,10,20,30},
            xtick={1,2,3,4,5},
            every axis plot/.append style={fill},
            nodes near coords={},
        ]
        \addplot [
            red!55!white!75!black,
            text=black,
            show sum on top,
        ] coordinates {
            (1,6)
            (2,7)
            (3,0)
            (4,0)
            (5,0)
        };
        \addplot [
            blue!55!white!75!black,
            text=black,
            show sum on top,
        ] coordinates {
            (1,0)
            (2,0)
            (3,10)
            (4,0)
            (5,0)
        };
        \addplot [
            green!55!white!75!black,
            text=black,
            show sum on top,
        ] coordinates {
            (1,0)
            (2,0)
            (3,0)
            (4,9)
            (5,3)
        };
        \end{axis}
    \end{tikzpicture}
    \subcaption{Medium.}
    \label{fig:q3medium}
    \end{subfigure}
    ~
    \begin{subfigure}[t]{.23\textwidth}
    \centering
    \begin{tikzpicture}[every node/.style={scale=0.8}]
        \begin{axis}
        [
            ylabel = Answers,
            xlabel = 5-point Likert Scale,
            ybar stacked,
            height=2.75cm,
            width=4.5cm,
            ymin = 0,
            ymax = 35,
            ytick={0,10,20,30},
            xtick={1,2,3,4,5},
            every axis plot/.append style={fill},
            nodes near coords={},
        ]
        \addplot [
            red!55!white!75!black,
            text=black,
            show sum on top,
        ] coordinates {
            (1,2)
            (2,7)
            (3,0)
            (4,0)
            (5,0)
        };
        \addplot [
            blue!55!white!75!black,
            text=black,
            show sum on top,
        ] coordinates {
            (1,0)
            (2,0)
            (3,10)
            (4,0)
            (5,0)
        };
        \addplot [
            green!55!white!75!black,
            text=black,
            show sum on top,
        ] coordinates {
            (1,0)
            (2,0)
            (3,0)
            (4,23)
            (5,16)
        };
        \end{axis}
    \end{tikzpicture}
    \subcaption{Hard.}
    \label{fig:q3hard}
    \end{subfigure}
    
    \caption{Bar charts of answers for question Q3 (``The challenge was just right (balance)'').}
    \label{fig:feedback_q3}
\end{figure*}

\begin{figure*}[!ht]
    \centering
    
    \begin{subfigure}[t]{.23\textwidth}
    \centering
    \begin{tikzpicture}[every node/.style={scale=0.8}]
        \begin{axis}
        [
            ylabel = Answers,
            xlabel = 5-point Likert Scale,
            ybar stacked,
            height=2.75cm,
            width=4.5cm,
            ymin = 0,
            ymax = 23,
            ytick={0,10,20},
            xtick={1,2,3,4,5},
            every axis plot/.append style={fill},
            nodes near coords={},
        ]
        \addplot [
            red!55!white!75!black,
            text=black,
            show sum on top,
        ] coordinates {
            (1,5)
            (2,5)
            (3,0)
            (4,0)
            (5,0)
        };
        \addplot [
            blue!55!white!75!black,
            text=black,
            show sum on top,
        ] coordinates {
            (1,0)
            (2,0)
            (3,2)
            (4,0)
            (5,0)
        };
        \addplot [
            green!55!white!75!black,
            text=black,
            show sum on top,
        ] coordinates {
            (1,0)
            (2,0)
            (3,0)
            (4,5)
            (5,1)
        };
        \end{axis}
    \end{tikzpicture}
    \subcaption{Easy.}
    \label{fig:crosseasy}
    \end{subfigure}
    ~
    \begin{subfigure}[t]{.23\textwidth}
    \centering
    \begin{tikzpicture}[every node/.style={scale=0.8}]
        \begin{axis}
        [
            ylabel = Answers,
            xlabel = 5-point Likert Scale,
            ybar stacked,
            height=2.75cm,
            width=4.5cm,
            ymin = 0,
            ymax = 23,
            ytick={0,10,20},
            xtick={1,2,3,4,5},
            every axis plot/.append style={fill},
            nodes near coords={},
        ]
        \addplot [
            red!55!white!75!black,
            text=black,
            show sum on top,
        ] coordinates {
            (1,4)
            (2,4)
            (3,0)
            (4,0)
            (5,0)
        };
        \addplot [
            blue!55!white!75!black,
            text=black,
            show sum on top,
        ] coordinates {
            (1,0)
            (2,0)
            (3,4)
            (4,0)
            (5,0)
        };
        \addplot [
            green!55!white!75!black,
            text=black,
            show sum on top,
        ] coordinates {
            (1,0)
            (2,0)
            (3,0)
            (4,5)
            (5,0)
        };
        \end{axis}
    \end{tikzpicture}
    \subcaption{Medium.}
    \label{fig:crossmedium}
    \end{subfigure}
    ~
    \begin{subfigure}[t]{.23\textwidth}
    \centering
    \begin{tikzpicture}[every node/.style={scale=0.8}]
        \begin{axis}
        [
            ylabel = Answers,
            xlabel = 5-point Likert Scale,
            ybar stacked,
            height=2.75cm,
            width=4.5cm,
            ymin = 0,
            ymax = 23,
            ytick={0,10,20},
            xtick={1,2,3,4,5},
            every axis plot/.append style={fill},
            nodes near coords={},
        ]
        \addplot [
            red!55!white!75!black,
            text=black,
            show sum on top,
        ] coordinates {
            (1,2)
            (2,4)
            (3,0)
            (4,0)
            (5,0)
        };
        \addplot [
            blue!55!white!75!black,
            text=black,
            show sum on top,
        ] coordinates {
            (1,0)
            (2,0)
            (3,7)
            (4,0)
            (5,0)
        };
        \addplot [
            green!55!white!75!black,
            text=black,
            show sum on top,
        ] coordinates {
            (1,0)
            (2,0)
            (3,0)
            (4,15)
            (5,11)
        };
        \end{axis}
    \end{tikzpicture}
    \subcaption{Hard.}
    \label{fig:crosshard}
    \end{subfigure}
    
    \caption{Bar charts of answers for question Q3 (``The challenge was just right'') of 43 players for 74 levels. These players answered they enjoy battles.}
    \label{fig:cross_feedback}
\end{figure*}
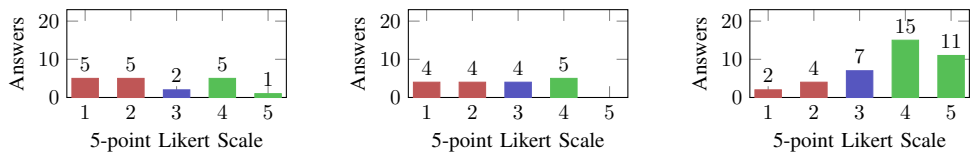

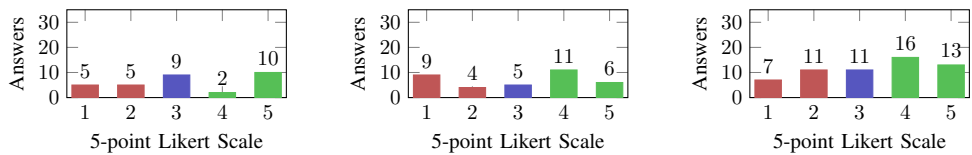
\begin{figure*}[!ht]
    \centering
    
    \begin{subfigure}[t]{.23\textwidth}
    \centering
    \begin{tikzpicture}[every node/.style={scale=0.8}]
        \begin{axis}
        [
            ylabel = Answers,
            xlabel = 5-point Likert Scale,
            ybar stacked,
            height=2.75cm,
            width=4.5cm,
            ymin = 0,
            ymax = 35,
            ytick={0,10,20,30},
            xtick={1,2,3,4,5},
            every axis plot/.append style={fill},
            nodes near coords={},
        ]
        \addplot [
            red!55!white!75!black,
            text=black,
            show sum on top,
        ] coordinates {
            (1,5)
            (2,5)
            (3,0)
            (4,0)
            (5,0)
        };
        \addplot [
            blue!55!white!75!black,
            text=black,
            show sum on top,
        ] coordinates {
            (1,0)
            (2,0)
            (3,9)
            (4,0)
            (5,0)
        };
        \addplot [
            green!55!white!75!black,
            text=black,
            show sum on top,
        ] coordinates {
            (1,0)
            (2,0)
            (3,0)
            (4,2)
            (5,10)
        };
        \end{axis}
    \end{tikzpicture}
    \subcaption{Easy.}
    \label{fig:q4easy}
    \end{subfigure}
    ~
    \begin{subfigure}[t]{.23\textwidth}
    \centering
    \begin{tikzpicture}[every node/.style={scale=0.8}]
        \begin{axis}
        [
            ylabel = Answers,
            xlabel = 5-point Likert Scale,
            ybar stacked,
            height=2.75cm,
            width=4.5cm,
            ymin = 0,
            ymax = 35,
            ytick={0,10,20,30},
            xtick={1,2,3,4,5},
            every axis plot/.append style={fill},
            nodes near coords={},
        ]
        \addplot [
            red!55!white!75!black,
            text=black,
            show sum on top,
        ] coordinates {
            (1,9)
            (2,4)
            (3,0)
            (4,0)
            (5,0)
        };
        \addplot [
            blue!55!white!75!black,
            text=black,
            show sum on top,
        ] coordinates {
            (1,0)
            (2,0)
            (3,5)
            (4,0)
            (5,0)
        };
        \addplot [
            green!55!white!75!black,
            text=black,
            show sum on top,
        ] coordinates {
            (1,0)
            (2,0)
            (3,0)
            (4,11)
            (5,6)
        };
        \end{axis}
    \end{tikzpicture}
    \subcaption{Medium.}
    \label{fig:q4medium}
    \end{subfigure}
    ~
    \begin{subfigure}[t]{.23\textwidth}
    \centering
    \begin{tikzpicture}[every node/.style={scale=0.8}]
        \begin{axis}
        [
            ylabel = Answers,
            xlabel = 5-point Likert Scale,
            ybar stacked,
            height=2.75cm,
            width=4.5cm,
            ymin = 0,
            ymax = 35,
            ytick={0,10,20,30},
            xtick={1,2,3,4,5},
            every axis plot/.append style={fill},
            nodes near coords={},
        ]
        \addplot [
            red!55!white!75!black,
            text=black,
            show sum on top,
        ] coordinates {
            (1,7)
            (2,11)
            (3,0)
            (4,0)
            (5,0)
        };
        \addplot [
            blue!55!white!75!black,
            text=black,
            show sum on top,
        ] coordinates {
            (1,0)
            (2,0)
            (3,11)
            (4,0)
            (5,0)
        };
        \addplot [
            green!55!white!75!black,
            text=black,
            show sum on top,
        ] coordinates {
            (1,0)
            (2,0)
            (3,0)
            (4,16)
            (5,13)
        };
        \end{axis}
    \end{tikzpicture}
    \subcaption{Hard.}
    \label{fig:q4hard}
    \end{subfigure}
    
    \caption{Bar charts of answers for question Q4 (``The enemies I faced were created by humans'').}
    \label{fig:feedback_q4}
\end{figure*}

A total of 96 players faced our enemies in the game, and 75 answered all the questions. They played 124 levels with enemies randomly placed in rooms. Regarding difficulty, they played 31 levels with easy enemies, 35 with medium enemies, and 58 with hard enemies. After playing the levels, the players answered how much agree or disagree, on a five-point Likert scale, with the following statements:

\begin{description}
\item[\textbf{Q1}] The level was fun to play;
\item[\textbf{Q2}] The enemies of this level were difficult to defeat;
\item[\textbf{Q3}] The challenge was just right (balance);
\item[\textbf{Q4}] The enemies I faced were created by humans.
\end{description}

In our experiments, we had 51 out of 75 players declaring to have considerable experience with games in general and 53 out of 75 with Action-Adventure games.
\autoref{fig:feedback} shows the overall results of players' feedback for each question of our questionnaire. The players enjoyed most the levels played (\autoref{fig:q1}), and \autoref{fig:q2} shows the players had no difficulty defeating the enemies in 80 levels. About the challenge suitability (\autoref{fig:q3}), they answered that the challenge was just right in half of the levels (62 out of 124) and neutral for 25 of the levels.
Finally, in \autoref{fig:q4}, we observe that the players sustained that humans created our enemies in 58 of the levels.

\autoref{fig:feedback_q1} shows results of question Q1 for each difficulty range. The players enjoyed most levels regardless of the difficulty of the enemies. However,  they preferred medium and hard levels since the negative answers decreased while positive ones increased.

\autoref{fig:feedback_q2} shows results of question Q2 for each difficulty.
In \autoref{fig:q2easy},  the players felt that the enemies in 20 out of 31 easy levels were not challenging to defeat, and there were only four levels with enemies perceived as harder to overcome. For the medium levels,  the players reported easy enemies to defeat in 15, hard in 11, and neutral in 9 levels (\autoref{fig:q2medium}). Finally, from 58 hard levels, \autoref{fig:q2hard} shows enemies hard to defeat in 29 levels based on players' feedback. They perceived enemies as easy to defeat in only 10 levels and neutral in 18 levels. Thus, the results about the difficulty range of our enemies corroborate our settings for difficulty values.

\autoref{fig:feedback_q3} shows results of question Q3 for each difficulty.
Figures \ref{fig:q3easy} and \ref{fig:q3medium} illustrate that approximately half of the players agree and half disagree that easy and medium levels challenges were just right. In contrast, \autoref{fig:q3hard} shows that the players perceived the challenge as adequate in most hard levels (39 out of 58).

Besides the questionnaire to evaluate levels, we also asked the players if they enjoyed battling during their gameplay. \autoref{fig:cross_feedback} shows the feedback of the players who confirmed such question. The answers are analogous to \autoref{fig:feedback_q3}, and these players also preferred the levels with harder enemies, as shown in \autoref{fig:crosshard}. Considering such results, we believe our levels probably could present better challenges if we mix up some enemies with different difficulty degrees.

\autoref{fig:feedback_q4} shows results of question Q4 for each difficulty.
The results indicate that players notice enemies as human-made in most levels regarding the difficulty degree of the enemies. Besides, this perception is more visible for medium and hard enemies than easy ones. These results mean that if the enemy is harder to overcome, the players perceive them as more carefully designed.

\section{Conclusion} \label{sec:conclusion}

In this paper, we introduced an illumination approach that extended the enemy generation presented by Pereira et al. \cite{ref:pereira2021procedural}. We advanced the previous method by illuminating the enemies through the MAP-Elites approach. Our experiments showed that the players had fun while playing the levels with the enemies generated regardless of the difficulty. The results corroborated the difficulty values we set to create easy, medium, and hard enemies. Furthermore, our enemies were carefully designed in enough way to be perceived as human-made enemies. As future works, we intend to extend our work with the Constrained MAP-Elites approach \cite{ref:gravina2019procedural}. With this approach, we can generate multiple enemies and avoid incoherent enemies.



\section*{Acknowledgments}

We acknowledge the financial support of the National Council for Scientific and Technological Development (CNPq).

\bibliographystyle{IEEEtran}
\bibliography{main}

\end{document}